\documentclass[conference]{IEEEtran}
\IEEEoverridecommandlockouts
\usepackage{array}
\usepackage[caption=false,font=normalsize,labelfont=sf,textfont=sf]{subfig}
\usepackage{textcomp}
\usepackage{stfloats}
\usepackage{url}
\usepackage{verbatim}
\usepackage{graphicx}
\usepackage{subcaption}
\usepackage{caption}
\usepackage{multirow}
\usepackage{pgfplots}
\usepackage{pgfplotstable}
\pgfplotsset{compat=1.16}
\usepackage[inkscapeformat=png]{svg}
\usepackage{adjustbox}
\usepackage{xcolor}

\usepackage{amsmath,amssymb,amsfonts}
\usepackage{textcomp}
\usepackage{xcolor}
\usepackage{enumitem}
\usepackage{booktabs}
\usepackage{colortbl}
\definecolor{lightgray}{gray}{0.9}
\usepackage{pifont}
\usepackage{algorithm}
\usepackage{algpseudocode}
\usepackage{makecell}
\usepackage{hyperref}
\usepackage{tabularx}
\newcolumntype{Y}{>{\centering\arraybackslash}X}

\makeatletter
\algrenewcommand\alglinenumber[0]{}
\makeatother

\def\BibTeX{{\rm B\kern-.05em{\sc i\kern-.025em b}\kern-.08em
    T\kern-.1667em\lower.7ex\hbox{E}\kern-.125emX}}

\begin{document}

\title{ROSA-RL: Uncertainty-Aware Roundabout Optimized Speed Advisory with Reinforcement Learning}

\author{\IEEEauthorblockN{Anna-Lena Schlamp\thanks{\copyright 2026 IEEE. Personal use of this material is permitted. Permission from IEEE must be obtained for all other uses, in any current or future media, including reprinting/republishing this material for advertising or promotional purposes, creating new collective works, for resale or redistribution to servers or lists, or reuse of any copyrighted component of this work in other works. This article has been accepted for publication in the proceedings of the \textit{2026 IEEE International Conference on Intelligent Transportation Systems (ITSC)}. This is the accepted manuscript version. The final version will be available in IEEE Xplore. DOI: To appear.}}
\IEEEauthorblockA{\textit{Institute AImotion Bavaria} \\
\textit{Technische Hochschule Ingolstadt}\\
Ingolstadt, Germany \\
anna-lena.schlamp@thi.de}
\and
\IEEEauthorblockN{Jeremias Gerner}
\IEEEauthorblockA{\textit{Institute AImotion Bavaria} \\
\textit{Technische Hochschule Ingolstadt}\\
Ingolstadt, Germany \\
jeremias.gerner@thi.de}
\and
\IEEEauthorblockN{Klaus Bogenberger}
\IEEEauthorblockA{\textit{School of Engineering and Design} \\
\textit{Technical University of Munich}\\
Munich, Germany \\
klaus.bogenberger@tum.de}
\and
\IEEEauthorblockN{Werner Huber}
\IEEEauthorblockA{\textit{CARISSMA Institute of Automated Driving
(C-IAD)} \\
\textit{Technische Hochschule Ingolstadt}\\
Ingolstadt, Germany \\
werner.huber@thi.de}
\and
\IEEEauthorblockN{Stefanie Schmidtner}
\IEEEauthorblockA{\textit{Institute AImotion Bavaria} \\
\textit{Technische Hochschule Ingolstadt}\\
Ingolstadt, Germany \\
stefanie.schmidtner@thi.de}
}

\maketitle

\begin{abstract}
Roundabouts challenge automated driving in mixed traffic, as heterogeneous and non-deterministic human behavior, unknown driving intentions, and high interaction complexity create uncertainty about whether the conflict zone will be blocked or available at the moment of entry.
We present ROSA-RL -- uncertainty-aware Roundabout Optimized Speed Advisory with Reinforcement Learning. It enables safe and efficient roundabout entry for automated and human-driven vehicles in mixed traffic through probabilistic conflict forecasting. A Transformer-based model predicts conflict zone occupancy over a five-second horizon, capturing multi-agent interactions to anticipate upcoming conflicts and available gaps. The prediction outputs encode uncertainty in future motion and intent, and augment the state of a classical RL framework, enabling uncertainty-aware speed coordination. Evaluated in simulations grounded in real-world data, ROSA-RL can effectively handle uncertainty and outperform a comparable model-based baseline, closing the gap to an ideal setting assuming fully known occupancy while improving traffic efficiency and safety. The source code of this work is available under: github.com/urbanAIthi/ROSA-RL.
\end{abstract}


\section{Introduction}
Roundabouts present a major challenge for automated driving, particularly in mixed traffic, as traffic flow and merging depend on mutual interactions among road users. Automated Vehicles (AVs) cannot rely on implicit communication cues and must instead anticipate the highly heterogeneous and non-deterministic behavior of human drivers under strict real-time constraints \cite{Frber2015KommunikationsproblemeZA, PredictingBehaviourNaveed2019}. A key challenge lies in intention prediction -- inferring whether road users will enter, yield, continue circulating, or exit the roundabout -- making the environment partially observable and often creating ambiguous situations that challenge safe coexistence \cite{PredictingBehaviourNaveed2019, Okumura2016ChallengesIP}. Conservative fallback strategies, such as stopping before entry, can ensure safety but often reduce traffic efficiency \cite{EmissionAnalysisChristofa2018}, while abrupt braking or delayed decisions may undermine trust and perceived safety for both human drivers and AV occupants \cite{Carlowitz2026BalancingCA, Stange2022PleaseSN}.
Proactive control is therefore essential to jointly address efficiency, safety, and acceptance in mixed traffic at roundabouts \cite{Carlowitz2026BalancingCA, Montanaro2018TowardsCA}. However, its effectiveness critically depends on robust handling of prediction uncertainty due to unknown driving intentions \cite{SchlampROSA25}. While model-based approaches define the strong performance baseline in deterministic, fully observable settings \cite{SchlampGLOSA23}, their optimality does not extend to uncertain traffic dynamics, where Reinforcement Learning (RL) can learn to act more robustly.
To address this, we introduce ROSA-RL, an RL–based Roundabout Optimized Speed Advisory (ROSA) system. By leveraging probabilistic conflict prediction, ROSA-RL enables vehicles to proactively adapt to evolving gaps in circulating traffic via uncertainty-aware speed coordination, ensuring safe and efficient roundabout entry.
\subsection{Vehicle Coordination at Roundabouts} \label{sec:coordination}
Several studies propose a coordination among automated vehicles to improve traffic flow and efficiency at roundabouts. Methods include vehicle clustering \cite{BiLevelBakibillah2021} and sequence determination \cite{OptControlZhao2017, TrajektorienplanerLong2022}, adjusting driving behavior such as speed or acceleration accordingly. Simulation results demonstrate reductions in waiting times, fuel consumption, and emissions under diverse traffic conditions \cite{BiLevelBakibillah2021, OptControlZhao2017}.
However, prior works assume full knowledge of driving intentions (e.g., planned turning \cite{TrajektorienplanerLong2022}) and cooperation among vehicles in fully automated settings, requiring full penetration and limiting applicability in mixed traffic~\cite{Montanaro2018TowardsCA, OptControlZhao2017}.
Schlamp et al. introduced ROSA \cite{SchlampROSA25}, Roundabout Optimized Speed Advisory, a non-cooperative speed optimization framework for mixed traffic, where partial observability naturally arises from unknown human driving intentions. ROSA employs interaction-aware trajectory prediction to issue proactive speed advisories up to five seconds before entry, reducing conflicts and delays in real-world traffic scenarios. However, conflicts are deterministically inferred from predicted trajectories via rule-based spatial checks, and speed advisories rely on basic kinematic models, thereby ignoring the inherent uncertainty in motion and intent prediction, which may limit performance.

\subsection{Traffic Prediction at Roundabouts}  
In interaction-based scenarios like roundabouts, proactive control of mixed traffic requires not only knowledge of the current dynamics but also anticipation of future conflicts and available gaps. A key challenge in roundabout traffic prediction is capturing inter-agent dependencies and road user intentions, which directly affect downstream tasks like coordinated speed guidance. Prior works on interaction modeling predominantly employ multi-agent prediction models that jointly encode all agents in the scene using a bird’s-eye view representation \cite{Scibior2021ImaginingTR, Westny2023MTPGOGP, Mann2022PredictingFO, Schreiber2019MotionEI}, and explicitly capture heterogeneous agent dynamics to enhance prediction accuracy \cite{SchlampROSA25, Zhang2024ExploringRD, Cheng2020ExploringDC}.
Architectures range from Recurrent Neural Networks (RNNs) and Graph Neural Networks (GNNs) to Transformer-based Neural Networks. RNNs are constrained by sequential processing and temporal context, while GNNs rely on handcrafted interaction graphs or map priors, limiting performance in highly dynamic scenes \cite{Westny2023MTPGOGP}. 
In contrast, Transformers offer a more flexible, data-driven framework with global context awareness and explicit interaction reasoning via self-attention, well suited for complex, interactive traffic scenarios like roundabouts \cite{Vaswani2017AttentionIA}. 

Building on these architectural foundations, existing methods can broadly be categorized by their prediction target into trajectory prediction and occupancy grid prediction.
Trajectory prediction at roundabouts typically uses three seconds of motion history to forecast five seconds ahead \cite{SchlampROSA25, Westny2023MTPGOGP}. Uncertainty in future motion and intent is commonly addressed through multimodal forecasting, producing a probability distribution over multiple plausible trajectories rather than a single deterministic path per agent \cite{Westny2023MTPGOGP, Liu2021MultimodalMP, Trentin2023LearningenabledMM}.
Occupancy prediction methods estimate the probability of future spatial occupancy in discretized grids \cite{PredictingBehaviourNaveed2019, Mohajerin2018MultiStepPO, Nadarajan2017PredictedoccupancyGF}, 
providing a compact, uncertainty-aware description of the traffic scene, ideal for downstream tasks like RL-based speed coordination. 
However, current methods are restricted to short prediction horizons ($\leq 3$s), coarse spatial resolution, and computationally intensive full-scene rasterization. Moreover, they remain largely unexplored in multi-agent roundabout scenarios. 
A conflict-zone–centric, data-driven occupancy prediction framework over a five-second horizon that jointly encodes all agents, their interactions, and uncertain intentions could address these limitations but has not yet been explored.

\subsection{Uncertainty-Aware Speed Optimization with RL}
Predicting future conflict zone occupancy at roundabouts is inherently uncertain due to the heterogeneous and non-deterministic behavior of human drivers and unknown driving intentions -- particularly whether they will exit before a given entry or continue circulating and block it. This uncertainty can be formally modeled as a Partially Observable Markov Decision Process (POMDP).
Reinforcement Learning provides a powerful framework for handling such uncertainty in sequential decision-making. 
Advanced approaches explicitly model uncertainty within the learning framework. For instance, Bayesian and belief-state RL use probability distributions over latent states or dynamics to support reasoning under partial observability \cite{Hu2025RiskAwareRL, Luis2024UncertaintyRI}. 
Distributional and risk‑sensitive RL extend classical value estimation by learning full return distributions or risk-augmented value functions, allowing policies to balance expected performance and risk in continuous control tasks \cite{Liu2023DistributionalRL, Xu2022DecisionmakingMO}.
While these methods improve robustness and exploration, their computational complexity can limit real-time applicability. As a computationally efficient alternative, recent works incorporate uncertainty implicitly within classical RL frameworks, e.g., by augmenting the state representation with risk-related features or shaping the reward with probabilistic cost terms \cite{Diehl2023UncertaintyAwareMO, Kahn2017UncertaintyAwareRL, AlShareeda2025WhenPH}. Notably, these lightweight extensions achieve robust decision-making in automated driving, providing a practical trade-off for applications like ROSA-RL.

In the context of vehicle speed optimization, RL has already been successfully applied at signalized intersections. In particular, Actor–Critic (AC) methods, which jointly learn a policy and value function, are able to discover optimal speed profiles that minimize stops and emissions \cite{SchlampGLOSA23, EcoWegener2021, EcoRLZhang2021}.
However, prior works are limited to deterministic settings with fully known traffic light phases. Under such conditions, model-based speed optimization derived from classical kinematics defines the strong performance baseline, which RL attains closely but does not exceed \cite{SchlampGLOSA23}. Given the uncertainty inherent in traffic prediction at roundabouts, model-based speed optimization no longer defines the strong baseline. In contrast, RL can learn to handle such uncertainty, but its effectiveness in this context has not been tested so far.
It remains an open question whether an RL-based speed advisory system for coordinating vehicles at roundabouts -- with shorter prediction horizons and higher dynamics than signalized intersections -- can (1) reach the strong performance baseline of a model-based approach in a deterministic, fully observable setting, and (2) better handle uncertainty to outperform model-based optimization under partial observability due to unknown driving intentions.

\subsection{Contributions}
We present ROSA-RL -- uncertainty-aware Roundabout Optimized Speed Advisory with Reinforcement Learning -- a real-time coordinated speed guidance for mixed traffic that explicitly accounts for prediction uncertainty. At its core, ROSA-RL introduces a data-driven prediction model to forecast conflict zone occupancy over a five-second horizon. 
The model captures all agents in the scene, their distinct dynamics and interactions, while avoiding the computational overhead of full-scene rasterization or agent-wise trajectory regression. 
The zone-centric occupancy predictions are used to encode uncertainty in future motion and intent, yielding a compact, probabilistic representation of upcoming conflicts and available gaps. 
Building on this, ROSA-RL employs uncertainty-aware Reinforcement Learning to generate proactive speed advisories for vehicles approaching and entering roundabouts, enabling smoother speed adaptation, reducing stops and delays, and improving efficiency and safety in mixed traffic without requiring cooperative behavior or full fleet penetration.
The main contributions of this work are:
\begin{itemize}
\item[1.] A probabilistic, conflict-zone–centric occupancy prediction model for roundabouts that captures multi-agent interactions and uncertain future behavior over a five-second horizon, with horizon-wise uncertainty quantification for downstream RL.
\item[2.] An uncertainty-aware, RL-based speed advisory framework that proactively coordinates vehicle speeds based on probabilistic conflict zone occupancy in mixed traffic.
\item[3.] A systematic comparison of model-based and RL-based speed optimization under full and partial observability, evaluating their effectiveness in handling prediction uncertainty and improving safety and efficiency in simulations based on real-world trajectory data.
\end{itemize}

\noindent
In Contribution 3, we examine: (1) whether RL-based speed optimization reaches the model-based performance baseline under deterministic, fully observable conditions with ground-truth zone occupancy; (2) the impact of prediction uncertainty and modality -- trajectory versus occupancy -- on speed advisory performance; and (3) since model-based optimality does not hold under prediction uncertainty, whether ROSA-RL can exploit probabilistic forecasts to better handle uncertainty and surpass model-based optimization in partially observable settings with unknown driving intentions.

ROSA-RL’s occupancy prediction is trained and validated on real-world trajectories from a typical urban single-lane roundabout (see Section \ref{dataset}). Its speed advisory framework is assessed in real-world traffic scenarios, modeling the observed occupancy in simulation. To support further research, we make the source code, trained models, and the simulation environment built on real-world data available under: github.com/urbanAIthi/ROSA-RL.

\begin{figure*}[tb]
    \centering
    \includegraphics[width=0.925\textwidth]{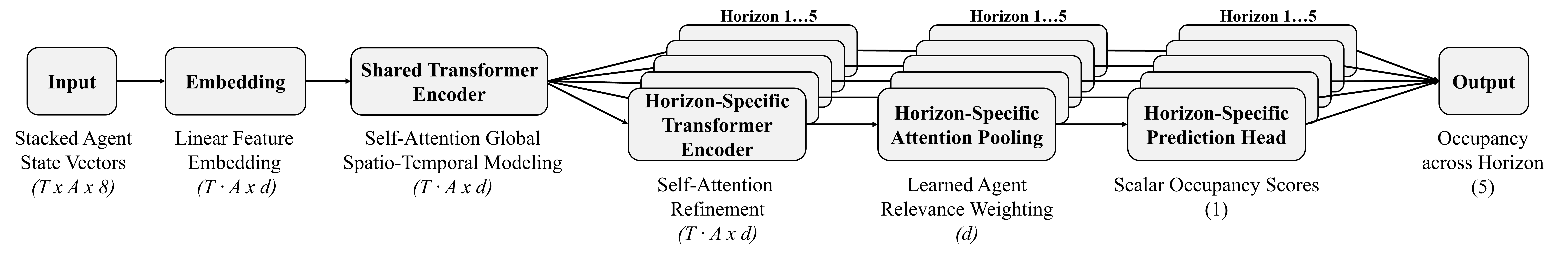}
    \caption{Architecture of the proposed zone-centric occupancy prediction model. The input sequence of agent features $(T \times A \times 8)$ is linearly embedded and processed by a shared Transformer encoder that captures spatio-temporal dependencies across all agents and timesteps, yielding contextualized token embeddings $(T \cdot A \times d)$. These are refined by five horizon-specific Transformer encoders, each followed by attention-based pooling that aggregates the token sequence into a compact scene-level representation $(d)$, and a dedicated MLP prediction head, yielding scalar occupancy scores $\hat{o}_1, \ldots, \hat{o}_5 \in \mathbb{R}$ over a five-second horizon.}  
    \label{fig:Architektur}
\end{figure*}

\section{Zone-centric Occupancy Prediction} \label{sec:prediction}
Occupancy prediction constitutes the foundation of ROSA-RL. We employ a Transformer architecture that captures inter-agent dependencies and traffic dynamics from raw trajectory data via self-attention. Adopting a conflict zone perspective, the model focuses on key areas of the roundabout while jointly encoding all agents in the scene. The resulting occupancy scores reflect uncertainty in future motion and intent, yielding a compact, probabilistic representation of upcoming conflicts and available gaps that supports downstream applications such as ROSA-RL's uncertainty-aware speed coordination.

\subsection{Dataset and Preprocessing} \label{dataset}
Several real-world datasets have been introduced to support traffic prediction at roundabouts. 
We use the \textit{openDD} dataset~\cite{Breuer2020openDDAL}, which provides 30~Hz drone-recorded trajectories of real-world roundabout traffic with detailed kinematic and semantic annotations, and focus on roundabout \textit{rdb1}, specifically the northern entry conflict zone. To reduce temporal redundancy and enable second-wise prediction, trajectories are downsampled to 1~Hz. As shown to improve accuracy \cite{SchlampROSA25}, input features are chosen to capture relevant agent dynamics: class label $c$, positional coordinates $p_x$ and $p_y$, velocity $v$, tangential and lateral accelerations \( a_{\text{tan}} \) and \( a_{\text{lat}} \), and the heading angle $\theta$.
The output targets are binary conflict zone occupancy labels over a five-second prediction horizon, with each one-second step labeled as occupied (1) if any road user is present and free (0) otherwise. 
The dataset is split into 80\% training, 10\% validation, and 10\% test sets, with balanced zone occupancy across splits.

\subsection{Model Architecture} \label{sec:model}
Our architecture builds on the multi-agent trajectory prediction model from ROSA~\cite{SchlampROSA25}, which employs a Transformer-based neural network~\cite{gerner2025FCO_TFCO}. 
The model adopts a holistic, scene-level view to capture interactions among road users within the unique structural dynamics of a roundabout. It is originally designed for single-step prediction of each agent’s immediate kinematic states using a three-second motion history, with autoregressive extension to multi-step forecasting. In this work, we enhance the architecture to directly predict conflict zone occupancy over a five-second horizon, while retaining the interaction-aware, multi-agent backbone from ROSA. 

\textbf{Input and Shared Encoder (adopted from ROSA):}
The model jointly processes the historical states of all interacting agents $A$ in the scene over the past $T=3$ seconds, each represented by a normalized feature vector  
\[
i = 
\left[
\text{c},\ 
p_x,\ p_y,\ 
v,\ 
a_{\text{tan}},\ a_{\text{lat}},\ 
\sin(\theta),\ \cos(\theta)\ 
\right] \in \mathbb{R}^{8}.
\]
The stacked input sequence $(T \times A \times 8)$ is flattened and linearly embedded into a latent space $(T \cdot A \times d)$.
The resulting agent–time tokens are processed by a shared Transformer encoder, whose self-attention explicitly captures spatio-temporal dependencies and interactions across the scene (see \cite{SchlampROSA25} for architectural details).

\textbf{Horizon-Specific Refinement and Prediction (specific to ROSA-RL):}
The shared encoder output, a sequence of contextualized token embeddings $(T \cdot A \times d)$, is processed by a set of five horizon-specific Transformer encoders, one per prediction step, enabling horizon-dependent refinement of spatio-temporal interactions. For each horizon, attention-based pooling aggregates the encoded tokens $(T \cdot A \times d)$ into a compact scene-level representation $(d)$, emphasizing agents that most influence scene dynamics for the respective forecast horizon. Dedicated prediction heads per horizon, each a lightweight Multilayer Perceptron (MLP), map the pooled embeddings to scalar occupancy scores $(1)$, with step-wise learnable parameters enabling adaptive confidence across short- and long-term forecasts (see Fig.~\ref{fig:Architektur}). 
Accordingly, the model outputs 
\[
\hat{p} = 
\left[ 
\hat{o}_1,\ 
\hat{o}_2,\ 
\hat{o}_3,\ 
\hat{o}_4,\ 
\hat{o}_5\ 
\right] \in \mathbb{R}^{5},
\]
where $\hat{o}_h$ denotes the predicted occupancy score for horizon~$h$. 
Binary cross-entropy is used as the training signal, with an additional logit regularization term to prevent overconfident outputs and stabilize training. 

\subsection{Evaluation Setup}
Zone-centric occupancy prediction is evaluated using standard binary classification metrics. Predictions are generated at one-second intervals over a five-second horizon and binarized with a decision threshold of $0.5$. We report F1-score, precision, recall, area under the receiver operating characteristic curve (ROC-AUC), and area under the precision--recall curve (PR-AUC). 
Precision measures true occupancy detections without producing false positives, while recall captures actual occupancies, minimizing missed events. F1 balances both, reflecting the trade-off between avoiding unnecessary speed advisories and maximizing optimization opportunities. ROC-AUC quantifies threshold-independent discrimination between occupied and free zones, while PR-AUC emphasizes precision under class imbalance.
ROSA-RL’s model, which directly predicts occupancy over a five-second horizon, is benchmarked against the ROSA baseline, where trajectories are forecast autoregressively in one-second steps and occupancy is deterministically inferred from predicted positions. Importantly, trajectory prediction in ROSA demonstrates state-of-the-art performance in interaction-aware forecasting of heterogeneous road user motion at roundabouts and substantially surpasses prior models \cite{SchlampROSA25}.
Both methods use identical inputs and conflict zone definitions to ensure a fair comparison. 
Evaluation is performed on the $\approx$6,600 samples from the test set, each representing an independent traffic scenario (1s resolution). We consider a single entry conflict zone of \textit{rdb1}, represented by five binary occupancy labels per sample, one for each step of the five-second prediction horizon. The dataset is naturally imbalanced, with occupied zones comprising roughly 15\% of test samples. 
Results are reported for a single training run. Experiments with five different random seeds produced comparable results (standard deviation $\leq 0.02$), indicating stable training behavior. 

\subsection{Results}
\begin{table*}[tb] 
\centering
\captionsetup{justification=centering}
\caption{Performance comparison of zone-centric occupancy prediction in ROSA-RL and the trajectory-based ROSA baseline. ROC-AUC and PR-AUC apply only to ROSA-RL, as ROSA derives occupancies deterministically from trajectories.}
\begin{tabular}{
>{\centering\arraybackslash}p{1.4cm}
>{\centering\arraybackslash}p{1.4cm}
>{\centering\arraybackslash}p{1.4cm}
>{\centering\arraybackslash}p{1.4cm}
>{\centering\arraybackslash}p{1.4cm}
>{\centering\arraybackslash}p{1.4cm}|
>{\centering\arraybackslash}p{1.4cm}
>{\centering\arraybackslash}p{1.4cm}
>{\centering\arraybackslash}p{1.4cm}}
\hline
\noalign{\vskip 4pt}
\multirow{2.7}{*}{\textbf{\shortstack{Prediction\\Horizon (s)}}} 
& \multicolumn{5}{c}{\textbf{ROSA-RL's Occupancy Prediction Model}} 
& \multicolumn{3}{c}{\textbf{Trajectory-based ROSA Baseline}} \\
\noalign{\vskip 2pt}
\cline{2-9}
\noalign{\vskip 4pt}
 & \textbf{F1} & \textbf{Precision} & \textbf{Recall} & \textbf{ROC-AUC} & \textbf{PR-AUC} & \textbf{F1} & \textbf{Precision} & \textbf{Recall} \\
\noalign{\vskip 2pt}
\hline
\noalign{\vskip 2pt}
\textbf{1} & 0.94 & 0.93 & 0.94 & 0.997 & 0.984 & 0.98 & 0.98 & 0.97 \\
\textbf{2} & 0.91 & 0.90 & 0.91 & 0.995 & 0.974 & 0.93 & 0.96 & 0.91 \\
\textbf{3} & 0.83 & 0.83 & 0.82 & 0.983 & 0.911 & 0.83 & 0.85 & 0.82 \\
\textbf{4} & 0.72 & 0.71 & 0.72 & 0.957 & 0.770 & 0.70 & 0.68 & 0.72 \\
\textbf{5} & 0.58 & 0.63 & 0.55 & 0.927 & 0.647 & 0.59 & 0.59 & 0.60 \\
\noalign{\vskip 2pt}
\hline
\end{tabular}
\label{tab:occupancy_vs_rosabase}
\end{table*}

The proposed zone-centric occupancy prediction attains strong discriminative performance despite pronounced class imbalance (see Table~\ref{tab:occupancy_vs_rosabase}). Short-term predictions achieve near-perfect separation between occupied and free conflict zones, with ROC-AUC of 0.997 and PR-AUC of 0.984 at one second. While performance gradually decreases with increasing horizon, the model remains robust up to five seconds, achieving a ROC-AUC of 0.927 and a PR-AUC of 0.647. This clearly outperforms the expected performance of a random classifier ($\approx 0.15$).
Recall decreases from 0.94 at one second to 0.55 at five seconds, reflecting increasing uncertainty over longer horizons, while precision remains comparatively stable, resulting in a gradual decline in F1 from 0.94 to 0.58. Accuracy stays high, but is influenced by the dataset’s natural imbalance.

In comparison, the trajectory-based ROSA baseline outperforms ROSA-RL's occupancy prediction (OP) at short horizons (F1: 0.98 vs. 0.94 at 1s; 0.93 vs. 0.91 at 2s), reflecting the reliability of deterministic trajectory extrapolation under low motion uncertainty. From mid- to long-term horizons, OP becomes increasingly competitive and surpasses the baseline at 4s (F1: 0.70 vs. 0.72). Unlike trajectory prediction in ROSA (TP), which suffers increasing false positives from autoregressive rollouts and hard geometric post-processing, OP preserves a more balanced precision--recall trade-off by abstracting agent interactions to the conflict zone level. Overall, zone-centric OP matches conflict assessment from state-of-the-art TP while offering greater robustness for mid- to long-term horizons, supporting downstream applications like proactive speed advisory. 


\subsection{Uncertainty Quantification}\label{sec:uncertainty}
\begin{figure}[tb]
    \centering
    \includegraphics[width=0.45\textwidth]{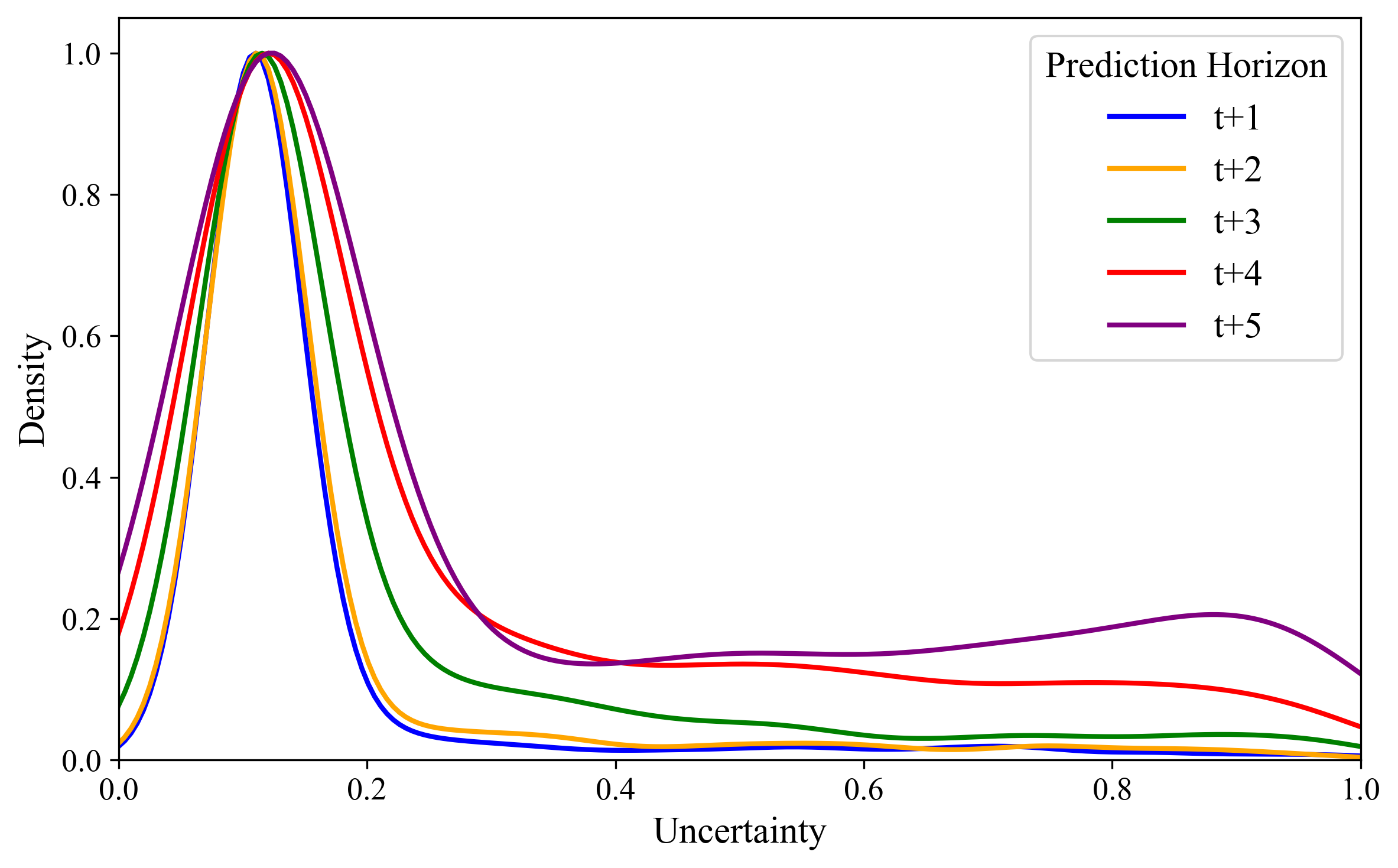}
    \caption{Density of scalar uncertainty across prediction horizons, increasing over time.}
    \label{fig:Uncertainty}
\end{figure}

Directly predicting occupancy scores allows a probabilistic interpretation of future conflict zone states. By retaining continuous outputs rather than enforcing binary decisions, uncertainty in future motion and road user intent can be derived from the signed distance between the predicted occupancy score $\hat{o}_h$ and the decision threshold $0.5$. Normalizing this distance to the range $[-1,1]$, we define a scalar uncertainty measure as the complement of its absolute value
\begin{equation}
    u = 1 - 2|\hat{o}_h - 0.5| \in [0,1],
\end{equation}
where values close to 1 indicate high uncertainty (scores near the threshold) and values close to 0 correspond to confident predictions (scores near 0 or 1).
Horizon-wise analysis of uncertainty distributions across the test set (see Fig.~\ref{fig:Uncertainty}) reveals a clear temporal structure. Short-term predictions exhibit sharply peaked, low-variance distributions, indicating high confidence. In contrast, longer horizons yield broader distributions with increased uncertainty, reflecting growing ambiguity in future traffic states. This structured evolution of uncertainty provides valuable confidence signals for downstream RL, enabling adaptive decision-making.

\section{Uncertainty-aware Roundabout Optimized Speed Advisory with Reinforcement Learning}
To address rising uncertainty in occupancy prediction and exploit confidence signals for decision-making, we introduce ROSA-RL, a Reinforcement Learning-based framework that generates uncertainty-aware speed advisories. Using five-second probabilistic forecasts, ROSA-RL augments a classical RL approach with occupancy and confidence information, effectively approximating the underlying POMDP as a belief-MDP with minimal computational overhead. This enables vehicles to proactively adapt to evolving gaps in circulating traffic, even under partial observability due to unknown human driving intentions.

\subsection{Gym Environment} \label{sec:Gym}
An RL agent learns to make decisions by interacting with its environment and maximizing rewards, i.e., feedback on its actions. To realize this closed-loop learning process, we employ the standardized Gymnasium interface \cite{Towers2024GymnasiumAS} coupled with the Simulation of Urban MObility (SUMO) \cite{behrisch2011sumo}, a microscopic traffic simulator. Within this environment, a single-lane roundabout approach is implemented, and real-world traffic scenes are replicated from the test split of the occupancy prediction dataset (see Section~\ref{dataset}) to model realistic vehicle interactions and conflict zone occupancy. This enables end-to-end training and evaluation under representative real-world traffic conditions. We consider a single connected and automated ego vehicle approaching the roundabout entry, which directly executes the issued speed advisories. Initially controlled by SUMO’s default vehicle dynamics, an episode starts once its time-to-arrival falls within the five-second prediction horizon ($\approx 42$~meters before the conflict zone). From this point onward, at each simulation step (1s resolution), ROSA-RL receives the current traffic state, predicts probabilistic occupancy scores, issues a speed advisory according to its policy, and obtains the corresponding reward along with the next state and termination signal. After the ego vehicle passes the conflict zone, control reverts to SUMO’s default dynamics and the episode terminates. This setup ensures efficient and reproducible RL training while also integrating the model-based ROSA baseline within the same environment. Consequently, RL-based and model-based approaches are evaluated under identical simulation and input conditions, enabling a fair and systematic comparison.

\subsection{RL Agent Design}
For the speed optimization task, we employ Proximal Policy Optimization (PPO)~\cite{Schulman2017ProximalPO, stable-baselines3}, an on-policy actor–critic algorithm well suited for continuous control problems. PPO iteratively updates a stochastic policy using a clipped surrogate objective, promoting stable learning while preserving effective exploration.
Both actor and critic are implemented as MLPs with two hidden layers of 64 neurons each. The policy outputs continuous actions in $[-1, 1]$, which are linearly scaled to the admissible speed range $[v_{\text{min}}, v_{\text{max}}]$ to generate speed advisories. 
The RL agent’s state, extracted from the SUMO simulation and provided as input to the neural network, comprises the current vehicle speed, 
the distance to the roundabout entry conflict zone, 
and the expected arrival times at the conflict zone based on the current and maximum speed. 
For each arrival time, the state further includes (i) the binarized occupancy prediction and (ii) the signed distance between the predicted occupancy score $\hat{o}_h$ and the decision threshold $0.5$, normalized to $[-1,1]$ (see Section \ref{sec:uncertainty}). This signed quantity encodes both the predicted occupancy via its sign and the associated confidence via its magnitude, thereby providing a continuous representation of prediction uncertainty.
A binary flag indicates whether the arrival time falls within the five-second prediction horizon, distinguishing between known and unknown occupancy states.
The agent's optimization goal is defined by a reward function, which quantifies the effectiveness of the chosen speed advisory in SUMO. The overall aim is to minimize travel time and energy use while ensuring smooth, conflict-free roundabout entry. To this end, we employ a linear combination of three reward components. The first component, $r_{\text{waiting}}$, penalizes the agent when the vehicle is stopped, directly targeting conflict avoidance and delay reduction. $r_{\text{diff}}$ penalizes deviations between consecutive speed advisories to promote smooth driving behavior, while $r_{\text{step}}$ applies a small per-step penalty to encourage higher speeds and minimize travel time.
Empirically determined weights of 5 for $r_{\text{waiting}}$, 0.5 for $r_{\text{diff}}$, and 1 for $r_{\text{step}}$ yield the per-step reward:

\begin{equation}
    r = 5 * r_{\text{waiting}} + 0.5 * r_{\text{diff}} + 1 * r_{\text{step}}.
\end{equation}

\subsection{Evaluation Scenarios}
In the test split of the occupancy prediction dataset, only 15\% of scenarios require a full stop due to an occupied conflict zone. Consequently, only those scenarios offer potential for optimization through coordinated speed guidance. 
We evaluate 200 randomly selected scenarios -- 100 optimizable and 100 non-optimizable -- while training the RL agent on a balanced subset of the remaining samples to assess generalization to unseen traffic situations. Results are consistent across five random seeds, demonstrating stable training. 
Simulation runs with and without speed advisories are compared to quantify efficiency and safety effects under real-world conditions. Efficiency is evaluated for both an Internal Combustion Engine (ICE\footnote{The standardized EURO4 vehicle model provided by SUMO is used.}) and a Battery Electric Vehicle (BEV), aggregating BEV energy consumption, ICE fuel consumption and CO$_2$ emissions, travel time, waiting time, and number of stops across scenarios. Safety effects are assessed indirectly via reductions in vehicle stops, serving as a proxy for smoother interactions, increased temporal separation and resolved conflicts.
To evaluate RL-based speed advisories at roundabouts, we conduct a structured comparison across fully and partially observable settings.
We analyze (1) MB-GT, model-based~(MB) optimization with ground-truth (GT) occupancy, and (2) RL-GT, its RL-based counterpart to assess whether RL can reach the model-based baseline under deterministic and fully observable conditions.
To isolate the impact of prediction modality and uncertainty, we compare (3) MB-TP, model-based ROSA with deterministic occupancy inference from predicted trajectories (see~\cite{SchlampROSA25}), and (4) MB-OP, model-based speed advisory operating on binarized occupancy prediction. To examine whether RL can exploit uncertainty under partial observability and outperform model-based optimization, we compare (5) RL-OP, which relies on binarized occupancy prediction without explicit uncertainty modeling, and (6)~UA-RL-OP, our proposed uncertainty-aware ROSA-RL leveraging probabilistic occupancy prediction, against model-based baselines.

\begin{table*}[ht]
\centering
\caption{Comparison of model- and RL-based speed optimization at roundabouts across observability and prediction modalities (GT vs. OP vs. TP), reporting average efficiency and safety gains in optimizable and non-optimizable scenarios.}
\label{tab:ROSA-RL_performance}
\setlength{\tabcolsep}{4.8pt}
\begin{tabular}{lcccccc|cccccc}
\hline 
\noalign{\vskip 4pt}
\multirow{3.7}{*}{\makecell[l]
{\textbf{Performance}\\\textbf{Metric}}} 
& \multicolumn{6}{c}{\textbf{Optimizable Scenarios}} 
& \multicolumn{6}{c}{\textbf{Non-Optimizable Scenarios}} \\
\noalign{\vskip 2pt}
\cline{2-13} 
\noalign{\vskip 4pt}
& \textbf{(1)} & \textbf{(2)} & \textbf{(3)} & \textbf{(4)} & \textbf{(5)} & \textbf{(6)}
& \textbf{(1)} & \textbf{(2)} & \textbf{(3)} & \textbf{(4)} & \textbf{(5)} & \textbf{(6)} \\
& \textbf{MB-GT} & \textbf{RL-GT} & \textbf{MB-TP} & \textbf{MB-OP} & \textbf{RL-OP} & \textbf{UA-RL-OP} 
& \textbf{MB-GT} & \textbf{RL-GT} & \textbf{MB-TP} & \textbf{MB-OP} & \textbf{RL-OP} & \textbf{UA-RL-OP} \\
\noalign{\vskip 2pt}
\hline 
\noalign{\vskip 4pt}
\makecell[l]{\textbf{BEV Energy}\\\textbf{Consumption}} 
& -61.3\% & -68.0\% & -56.0\% & -53.1\% & -47.4\% & -85.6\% 
& -2.3\% & -2.6\% & -13.4\% & -5.0\% & -6.1\% & -2.4\% \\[8pt]

\makecell[l]{\textbf{ICE Fuel \&}\\\textbf{CO$_2$ Emissions}} 
& -35.2\% & -16.3\% & -28.0\% & -24.9\% & -13.5\% & -17.5\% 
& +0.2\% & +0.5\% & +8.1\% & +1.6\% & +2.4\% & +4.8\% \\[8pt]

\makecell[l]{\textbf{Travel}\\\textbf{Time}} 
& -9.2\% & -9.2\% & -5.8\% & -3.1\% & -5.3\% & -5.9\% 
& $\pm$0.0\% & +0.5\% & +7.0\% & +2.2\% & +2.5\% & +7.5\% \\[8pt]

\makecell[l]{\textbf{Waiting}\\\textbf{Time}} 
& -91.3\% & -90.7\% & -78.9\% & -62.3\% & -69.2\% & -89.9\% 
& $\pm$0.0\% & $\pm$0.0\% & +3.0\% & +1.0\% & +2.0\% & $\pm$0.0\% \\[8pt]

\makecell[l]{\textbf{Number of}\\\textbf{Stops}} 
& -78.0\% & -80.0\% & -60.0\% & -42.0\% & -52.0\% & -80.0\% 
& $\pm$0.0\% & $\pm$0.0\% & +3.0\% & +1.0\% & +2.0\% & $\pm$0.0\% \\
\noalign{\vskip 4pt}
\hline
\end{tabular}
\end{table*}

\subsection{Results}
Under fully observable conditions, (2) RL-GT achieves performance comparable to the model-based strong baseline (1) MB-GT across both optimizable and non-optimizable scenarios (see Table \ref{tab:ROSA-RL_performance}), demonstrating that RL can learn effective speed profiles to reach gaps in highly dynamic roundabout traffic. The observed degradation in ICE emissions can be attributed to slightly oscillatory speed advisories that induce short-term acceleration fluctuations, while BEV energy consumption is less affected.

Comparing model-based speed advisory across prediction modalities -- trajectory prediction ((3) MB-TP) and binarized occupancy prediction ((4) MB-OP) -- reveals a clear trade-off: In non-optimizable scenarios, MB-OP outperforms MB-TP by reducing false-positive conflict detections and avoiding unnecessary speed interventions. In optimizable scenarios, where safety- and efficiency-oriented speed guidance is most relevant, MB-TP achieves stronger improvements, particularly in stop reductions, indicating enhanced safety. However, MB-OP generates smoother speed advisories with fewer oscillations, leading to more stable emission outcomes. In summary, OP increases robustness under uncertainty, whereas TP achieves higher gains when optimization potential exists.

A comparison of model-based and RL-based speed guidance under uncertain binary occupancy predictions ((4) MB-OP vs. (5) RL-OP) shows that RL again achieves performance comparable to the model-based approach under identical inputs, with only a minor drawback in emissions. While incorporating uncertainty into model-based advisory is non-trivial, RL naturally enables policy learning from probabilistic state information.
This advantage becomes evident in (6) UA-RL-OP, our proposed uncertainty-aware ROSA-RL. Under partial observability, ROSA-RL consistently outperforms all other optimization variants, including the strong model-based ROSA baseline with state-of-the-art TP ((3) MB-TP). In optimizable scenarios, UA-RL-OP achieves substantially stronger reductions in waiting time (-90\% vs. -79\%) and stops (-80\% vs. -60\%), indicating more effective conflict resolution and improved traffic flow compared to MB-TP, while compensating for the performance loss associated with occupancy prediction. As with other RL variants, a minor emission drawback is observed (-18\% vs. -28\%) due to residual speed oscillations, whereas BEV energy consumption improves markedly (-86\% vs. -56\%). In non-optimizable scenarios, ROSA-RL remains robust and does not introduce additional stops compared to other methods, demonstrating resilience to false-positive detections. Overall, ROSA-RL effectively exploits prediction uncertainty to narrow the gap to fully observable baselines assuming GT occupancy (MB-GT and RL-GT), despite operating under partial observability arising from unknown driving intentions.

\section{Discussion and Conclusion}

Interaction-rich roundabout scenarios challenge automated driving in mixed traffic, as heterogeneous human behavior and unknown driving intentions introduce uncertainty about gap availability, making the environment partially observable. We propose ROSA-RL, an uncertainty-aware RL-based Roundabout Optimized Speed Advisory that enables vehicles to proactively adapt to evolving gaps in circulating traffic. Central to ROSA-RL is a Transformer model that predicts probabilistic conflict zone occupancy over a five-second horizon, capturing multi-agent interactions while avoiding full-scene rasterization or agent-wise trajectory regression. The model obtains strong short-term performance and remains robust over five seconds, achieving a more balanced precision--recall trade-off compared to conflict inference from state-of-the-art trajectory prediction. 
By encoding uncertainty in future motion and intent, zone-centric occupancy predictions provide a compact, probabilistic representation of upcoming conflicts and available gaps, which ROSA-RL integrates into a classical RL framework to approximate the underlying POMDP as a belief-MDP. 
Evaluations in real-world simulations show that an RL agent can learn effective speed profiles at roundabouts, achieving performance comparable to model-based optimization under full observability with ground-truth occupancy, the strong baseline in this setting. Under partial observability, ROSA-RL leverages prediction uncertainty to outperform model-based speed advisories, approaching fully observable performance while significantly reducing conflicts, delays, and emissions. 

To conclude, ROSA-RL effectively enhances traffic flow, efficiency, and safety in mixed traffic without requiring full fleet penetration, while also supporting perceived safety and acceptance for both human drivers and AV occupants. 
ROSA-RL is not limited to AVs but also extends to Connected Vehicles~(CVs) with human drivers. Moreover, its data-driven  architecture naturally scales to heterogeneous road users and multiple conflict zones.
It also generalizes to different roundabout layouts and even unsignalized intersections, given sufficiently diverse training data. Practical deployment assumes a central unit aggregating past and current trajectory data from roadside sensing infrastructure or CVs that continuously broadcast their local perception via Vehicle-to-Everything~(V2X) communication, acting as so-called Floating Car Observers (FCOs) \cite{gerner2023FCO}. Occupancy prediction can be performed either centrally or decentrally on the vehicles themselves using V2X data.
Future work may extend ROSA-RL toward cooperative multi-vehicle control and assess its capability to handle high-dimensional decision spaces. Occupancy prediction could be augmented with explicit intention estimation to assess whether maneuver-level foresight improves predictive accuracy. Moreover, robustness under varying traffic demands, the impact of different perception modalities, V2X-related uncertainty and latency, and human compliance warrant systematic investigation. Real-world experiments will be essential to validate our findings and support practical deployment.

\section*{Acknowledgement}
During manuscript preparation, ChatGPT was used for minor language and grammar refinement. All content was reviewed and edited by the authors, who take full responsibility. 
\newline


\begin{thebibliography}{10}

\bibitem{Frber2015KommunikationsproblemeZA}
Berthold Färber. “Kommunikationsprobleme zwischen autonomen Fahrzeugen und menschlichen Fahrern”. 
\newblock In: {\em Autonomes Fahren. Springer Berlin Heidelberg}. 2015. $DOI:~10.1007/978-3-662-45854-9-7$.

\bibitem{PredictingBehaviourNaveed2019}
Muhammad Naveed and Bj{\"o}rn {\AA}strand. “Predicting Agent Behaviour and State for Applications in a Roundabout Scenario Autonomous Driving”. 
\newblock In: {\em  Sensors} (2019). $DOI:~10.3390/s19194279$.

\bibitem{Okumura2016ChallengesIP}
Bunyo Okumura et al. “Challenges in Perception and Decision Making for Intelligent Automotive Vehicles: A Case Study”. 
\newblock In: {\em IEEE Transactions on Intelligent Vehicles} (2016), pp. 20–32. $DOI:~10.1109/TIV.2016.2551545$.

\bibitem{EmissionAnalysisChristofa2018}
Eleni Christofa and Michael Knodler. "Operational and Emission Analyses of Roundabouts Under Varied Vehicle and Pedestrian Demands". 2018.

\bibitem{Carlowitz2026BalancingCA}
Stefanie Carlowitz et al. “Balancing comfort and efficiency: Optimal deceleration rates at crosswalks and intersections in automated driving”. 
\newblock In: {\em Transportation Research Part F: Traffic Psychology and Behaviour} (2026). $DOI:~10.1016/j.trf.2025.06.029$.

\bibitem{Stange2022PleaseSN}
Vanessa Stange et al. “Please stop now, automated vehicle! – Passengers aim to avoid risk experiences in interactions with a crossing vulnerable road user at an urban junction”.
\newblock In: {\em Transportation Research Part F: Traffic Psychology and Behaviour} (2022). $DOI:~10.1016/j.trf.2022.04.001$.

\bibitem{Montanaro2018TowardsCA}
Umberto Montanaro et al. “Towards Connected Autonomous Driving: Review of Use-Cases”.
\newblock In: {\em Vehicle System Dynamics} (2018). $DOI:~10.1080/ 00423114.2018.1492142$.

\bibitem{SchlampROSA25}
Anna-Lena Schlamp et al. “ROSA: Roundabout Optimized Speed Advisory with Multi-Agent Trajectory Prediction in Multimodal Traffic”. 
\newblock In: {\em 2025 IEEE 28th International Conference on Intelligent Transportation Systems (ITSC)} (2025). $DOI:~10.1109/ITSC60802.2025.11423540$.

\bibitem{SchlampGLOSA23}
Anna-Lena Schlamp et al. “User-Centric Green Light Optimized Speed Advisory with Reinforcement Learning”. 
\newblock In: {\em 2023 IEEE 26th International Conference on Intelligent Transportation Systems (ITSC)} (2023). $DOI:~10.1109/ITSC57777.2023.10422501$.

\bibitem{BiLevelBakibillah2021}
A. S. M. Bakibillah et al. “Bi-Level Coordinated Merging of Connected and Automated Vehicles at Roundabouts”.  
\newblock In: {\em Sensors} (2021). $DOI:~10.3390/s21196533$.

\bibitem{OptControlZhao2017}
Liuhui Zhao, Andreas Malikopoulos, and Jackeline Rios Torres. “Optimal Control of Connected and Automated Vehicles at Roundabouts: An Investigation in a Mixed-Traffic Environment”. 
\newblock In: {\em 15th IFAC Symposium on Control in Transportation Systems} (2018). $DOI:~10.1016/j.ifacol.2018.07.013$.

\bibitem{TrajektorienplanerLong2022}
Keke Long et al. “Optimization based trajectory planner for multilane roundabouts with connected automation”. 
\newblock In: {\em Journal of Intelligent Transportation Systems} (2022). $DOI:~10.1080/15472450.2022.2046473$.

\bibitem{Scibior2021ImaginingTR}
Adam Scibior et al. "Imagining The Road Ahead: Multi-Agent Trajectory Prediction via Differentiable Simulation". 
\newblock In: {\em 2021 IEEE International Intelligent Transportation Systems Conference (ITSC)} (2021). $DOI:~10.1109/itsc48978.2021.9565113$.

\bibitem{Westny2023MTPGOGP}
Theodor Westny et al. "MTP-GO: Graph-Based Probabilistic Multi-Agent Trajectory Prediction With Neural ODEs". 
\newblock In: {\em IEEE Transactions on Intelligent Vehicles} (2023). $DOI:~10.1109/TIV.2023.3282308$.

\bibitem{Mann2022PredictingFO}
Khushdeep Singh Mann et al. “Predicting Future Occupancy Grids in Dynamic Environment with Spatio-Temporal Learning”.
\newblock In: {\em 2022 IEEE Intelligent Vehicles Symposium (IV)} (2022). $DOI:~10.48550/arXiv.2205.03212$.

\bibitem{Schreiber2019MotionEI}
Marcel Schreiber et al. “Motion Estimation in Occupancy Grid Maps in Stationary Settings Using Recurrent Neural Networks”.
\newblock In: {\em 2020 IEEE International Conference on Robotics and Automation (ICRA)} (2019). $DOI:~10.1109/ICRA40945.2020.9196702$.

\bibitem{Zhang2024ExploringRD}
Zikai Zhang. "Exploring rounD Dataset for Domain Generalization in Autonomous Vehicle Trajectory Prediction". 
\newblock In: {\em Sensors (Basel, Switzerland)} (2024). $DOI:~10.3390/s24237538$.

\bibitem{Cheng2020ExploringDC}
Hao Cheng et al. "Exploring Dynamic Context for Multi-path Trajectory Prediction". 
\newblock In: {\em 2021 IEEE International Conference on Robotics and Automation (ICRA)} (2020). $DOI:~10.1109/ICRA48506.2021.9562034$.

\bibitem{Vaswani2017AttentionIA}
Ashish Vaswani et al. "Attention is All you Need". 
\newblock In: {\em Neural Information Processing Systems (NIPS)} (2017).

\bibitem{Liu2021MultimodalMP}
Yicheng Liu et al. "Multimodal Motion Prediction with Stacked Transformers". 
\newblock In: {\em 2021 IEEE/CVF Conference on Computer Vision and Pattern Recognition (CVPR)} (2021). $DOI:~10.1109/CVPR46437.2021.00749$.

\bibitem{Trentin2023LearningenabledMM}
Vinicius Trentin et al. "Learning-enabled multi-modal motion prediction in urban environments". 
\newblock In: {\em 2023 IEEE Intelligent Vehicles Symposium (IV)} (2023). $DOI:~10.1109/IV55152.2023.10186684$.

\bibitem{Mohajerin2018MultiStepPO}
Nima Mohajerin and Mohsen Rohani. “Multi-Step Prediction of Occupancy Grid Maps With Recurrent Neural Networks”.
\newblock In: {\em 2019 IEEE/CVF Conference on Computer Vision and Pattern Recognition (CVPR)} (2018). $DOI:~10.1109/CVPR.2019.01085$.

\bibitem{Nadarajan2017PredictedoccupancyGF}
Parthasarathy Nadarajan et al. “Predicted-occupancy grids for vehicle safety applications based on autoencoders and the Random Forest algorithm”. 
\newblock In: {\em 2017 International Joint Conference on Neural Networks (IJCNN)} (2017). $DOI:~10.1109/IJCNN.2017.7965995$.

\bibitem{Hu2025RiskAwareRL}
Chuanping Hu et al. “Risk-Aware Reinforcement Learning for Non-Conservative Motion Planning in Uncertain Autonomous Driving Environments”. 
\newblock In: {\em IEEE Transactions on Intelligent Transportation Systems} (2025). $DOI:~10.1109/TITS.2025.35992606$.

\bibitem{Luis2024UncertaintyRI}
Carlos E. Luis et al. “Uncertainty Representations in State-Space Layers for Deep  Reinforcement Learning under Partial Observability”. 
\newblock In: {\em Trans. Mach. Learn. Res.} (2025). $DOI:~10.48550/arXiv.2409.16824$.

\bibitem{Liu2023DistributionalRL}
Qi Liu et al. “Distributional reinforcement learning with epistemic and aleatoric uncertainty estimation”. 
\newblock In: {\em Inf. Sci} (2023). $DOI:~10.1016/j.ins.2023.119217$.

\bibitem{Xu2022DecisionmakingMO}
Shuyuan Xu et al. “Decision-making Models on Perceptual Uncertainty with Distributional Reinforcement Learning”. 
\newblock In: {\em Green Energy and Intelligent Transportation} (2022). $DOI:~10.1016/j.geits.2022.100062$.

\bibitem{Diehl2023UncertaintyAwareMO}
Christopher Diehl et al. “Uncertainty-Aware Model-Based Offline Reinforcement Learning for Automated Driving”.
\newblock In: {\em IEEE Robotics and Automation Letters} (2023). $DOI:~10.1109/LRA.2023.3236579$.

\bibitem{Kahn2017UncertaintyAwareRL}
Gregory Kahn et al. “Uncertainty-Aware Reinforcement Learning for Collision Avoidance”.
\newblock In: {\em ArXiv} (2017).

\bibitem{AlShareeda2025WhenPH}
Sarah Al-Shareeda et al. “When Pedestrians Hesitate: PPO-Based RL Collision Avoidance in Uncertain Scenarios”. 
\newblock In: {\em 2025 International Conference on Smart Applications, Communications and Networking (SmartNets)} (2025). $DOI:~10.1109/SmartNets65254.2025.11106806$.

\bibitem{EcoWegener2021}
Marius Wegener et al. “Automated eco-driving in urban scenarios using deep reinforcement learning”. 
\newblock In: {\em Transportation Research Part C Emerging Technologies} (2021). $DOI:~10.1016/j.trc.2021.102967$.

\bibitem{EcoRLZhang2021}
Xiaoliang Zhang et al. “Eco-driving for Intelligent Electric Vehicles at Signalized Intersection: A Proximal Policy Optimization Approach”. 
\newblock In: {\em 6th International Conference on Information Science, Computer Technology and Transportation (ISCTT)} (2021).

\bibitem{Breuer2020openDDAL}
Antonia Breuer et al. "openDD: A Large-Scale Roundabout Drone Dataset". 
\newblock In: {\em 2020 IEEE 23rd International Conference on Intelligent Transportation Systems (ITSC)} (2020). $DOI:~10.1109/ITSC45102.2020.9294301$.

\bibitem{gerner2025FCO_TFCO}
Jeremias Gerner and Klaus Bogenberger and Stefanie Schmidtner. "Floating Car Observers in Intelligent Transportation Systems: Detection Modeling and Temporal Insights". 
\newblock In: {\em ArXiv} (2025). 

\bibitem{Towers2024GymnasiumAS}
Mark Towers et al. “Gymnasium: A Standard Interface for Reinforcement Learning Environments”. 
\newblock In: {\em ArXiv} (2024). $DOI:~10.48550/arXiv.2407.17032$.

\bibitem{behrisch2011sumo}
Michael Behrisch et al. “SUMO- Simulation of Urban MObility: An Overview”. 
\newblock In: {\em The Third International Conference on Advances in System Simulation} (2011).

\bibitem{Schulman2017ProximalPO}
John Schulman et al. “Proximal Policy Optimization Algorithms”. 
\newblock In: {\em ArXiv} (2017). 

\bibitem{stable-baselines3}
Antonin Raffin et al. “Stable-Baselines3: Reliable Reinforcement Learning Implementations”. \newblock In: {Journal of Machine Learning Research} (2021).

\bibitem{gerner2023FCO}
Jeremias Gerner et al. "Enhancing Realistic Floating Car Observers in Microscopic Traffic Simulation". 
\newblock In: {\em 2023 IEEE 26th International Conference on Intelligent Transportation Systems (ITSC)} (2023). $DOI:~10.1109/ITSC57777.2023.10422398$.


\end{thebibliography}
\end{document}